\documentclass[10pt, a4paper, fleqn]{article}
\usepackage{lrec2006}
\usepackage{graphicx}
\usepackage{url}
\expandafter\def\expandafter\UrlBreaks\expandafter{\UrlBreaks
  \do\a\do\b\do\c\do\d\do\e\do\f\do\g\do\h\do\i\do\j%
  \do\k\do\l\do\m\do\n\do\o\do\p\do\q\do\r\do\s\do\t%
  \do\u\do\v\do\w\do\x\do\y\do\z\do\A\do\B\do\C\do\D%
  \do\E\do\F\do\G\do\H\do\I\do\J\do\K\do\L\do\M\do\N%
  \do\O\do\P\do\Q\do\R\do\S\do\T\do\U\do\V\do\W\do\X%
  \do\Y\do\Z}

\usepackage{hyperref}
\usepackage{amsmath}

\usepackage{array}
\newcolumntype{P}[1]{>{\centering\arraybackslash}p{#1}}

\title{Monitoring Energy Trends through Automatic Information Extraction}

\name{Dilek K\"u\c{c}\"uk}

\address{T\"UB\.ITAK Energy Institute\\
Ankara{--}Turkey\\
dilek.kucuk@tubitak.gov.tr\\}

\abstract{Energy research is of crucial public importance but the use of computer science technologies like automatic text processing and data management for the energy domain is still rare. Employing these technologies in the energy domain will be a significant contribution to the interdisciplinary topic of ``energy informatics", just like the related progress within the interdisciplinary area of ``bioinformatics". In this paper, we present the architecture of a Web-based semantic system called EneMonIE (Energy Monitoring through Information Extraction) for monitoring up-to-date energy trends through the use of automatic, continuous, and guided information extraction from diverse types of media available on the Web. The types of media handled by the system will include online news articles, social media texts, online news videos, and open-access scholarly papers and technical reports as well as various numeric energy data made publicly available by energy organizations. The system will utilize and contribute to the energy-related ontologies and its ultimate form will comprise components for (i) text categorization, (ii) named entity recognition, (iii) temporal expression extraction, (iv) event extraction, (v) social network construction, (vi) sentiment analysis, (vii) information fusion and summarization, (viii) media interlinking, and (ix) Web-based information retrieval and visualization. Wits its diverse data sources, automatic text processing capabilities, and presentation facilities open for public use; EneMonIE will be an important source of distilled and concise information for decision-makers including energy generation, transmission, and distribution system operators, energy research centres, related investors and entrepreneurs as well as for academicians, students, other individuals interested in the pace of energy events and technologies.\\ \newline }

\begin{document}

\maketitleabstract

\section{Introduction}\label{sec:intro}

Energy is so an indispensable asset today that we cannot even think of being deprived of it. It is one of the important topics that influence both individuals and governments to great extents. Therefore, results of the research and trends on conventional and sustainable/renewable energy resources; on energy production, transmission, distribution, conversion, utilization, and forecast as well as that of recent research topics like energy efficiency and smart electricity grids have important implications on the society in general. To facilitate public access to recent energy trends, events, and research topics; the resources, algorithms, and tools offered within the discipline of computer science and information technologies can be customized, redesigned, and employed and this will be an important contribution to the interdisciplinary area of ``energy informatics".\\

In this paper, we propose a Web-based semantic system called EneMonIE (Energy Monitoring through Information Extraction) which facilitates monitoring up-to-date energy trends, events, and related energy research through the use of automatic, continuous, and guided (targeted) information extraction from diverse media available on the Internet.\\

After daily polling, filtering, and collection of data from these data sources, the system will apply several information extraction techniques on the collected data to extract and distil semantic information for indexing these media. Upon the completion of the information extraction procedure, the system will align, cluster, fuse, and summarize the collected data based on the extracted semantic information to arrive at up-to-date energy trends and research topics in addition to energy data. Furthermore, the system will enrich the social media texts by associating with them automatically determined opinions, through its sentiment analysis component.\\

The types of media to be handled by the system include:

\begin{enumerate}
  \item News articles available online
  \item Social media texts like tweets and blog posts
  \item News videos available online
  \item Open-access papers in scholarly journals, openly available academic theses, and technical reports available on preprint servers
  \item Open numeric and textual data publicly provided by energy-related organizations
\end{enumerate}

EneMonIE will have a modular and extensible architecture with pluggable information extraction and other text processing components. Thereby, if need be, upon the completion of EneMonIE, newer information extraction tools can be integrated into the system without the need for considerable effort, provided that these tools comply with the corresponding software interfaces of the overall system.\\

To the best of our knowledge, no semantic system similar to EneMonIE has been described in the literature. Yet, there are systems that have some common features to EneMonIE: the fully-automated and semi-automated systems described in \cite{kuccuk2011exploiting} and \cite{kuccuk2013semi} have utilized information extraction techniques to facilitate semantic news video retrieval. Within these systems news videos and news articles are processed, yet other types of media within the scope of EneMonIE, such as social media, are not considered. Another difference between EneMonIE and the aforementioned two systems is that, these systems are also proposed for the generic domain of news. Additionally, these systems lack other important text processing components like those for sentiment analysis and summarization while EneMonIE will comprise these components to enrich its output and presentation.\\

We should also note the existence of related websites like OpenEI\footnote{\url{https://openei.org}}, reegle\footnote{\url{https://www.reeep.org/reegleinfo}}, and SETIS\footnote{\url{https://setis.ec.europa.eu/}}. OpenEI is a website hosting its collaboratively-managed energy-related resources and energy data sets which aims to serve this data as linked open data to public use. Reegle, on the other hand, is a clean energy information portal enabling its users to search for clean energy and energy efficiency information that it includes. Finally, SETIS (Strategic Energy Technologies Information System) is a portal for hosting studies and reports for energy-related decision makers, particularly to help implement and monitor the European Strategic Energy Technology Plan (SET-Plan) in European Union countries. These three systems have features mainly related EneMonIE's feature on open energy and energy-related documents and news, but EneMonIE's main contribution of semantic processing of the ubiquitous, external, and independent diverse media is not available through these systems. Lastly, in \cite{woon2014forecasting} keyword-based methods have been utilized on scholarly articles to determine the research trends in renewable energy papers as keyword taxonomies. The main contribution of this paper is similar to a feature of EneMonIE on scholarly articles which will be handled by its text categorization component. Hence, EneMonIE will be a larger full-fledged system having –among others- this feature as well.

\begin{figure*}[h!]
\center \scalebox{0.525}
{\fbox{\includegraphics{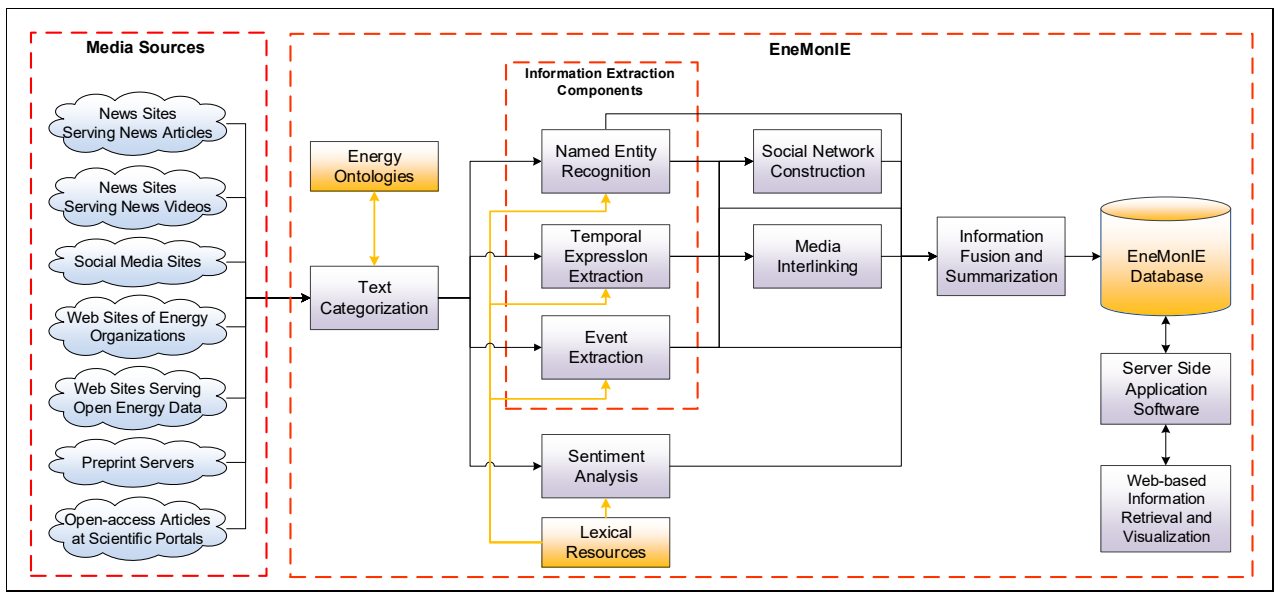}}}\caption{The EneMonIE Architecture}\label{fig:enemonie}
\end{figure*}

\section{EneMonIE Architecture}\label{sec:architecture}

The pipelined architecture of EneMonIE is demonstrated in Figure \ref{fig:enemonie} together with its data (media) sources. Functional components and resources (database, ontologies, and lexical resources) are shown with different colours to facilitate the distinction. For illustrative purposes, all information extraction components (named entity recognizer, temporal expression extractor, and event extractor) are shown as a single component, under the name of Information Extraction.\\

The main information extraction components of EneMonIE will be:
\begin{enumerate}
  \item \emph{Named entity recognition}: This task is usually defined as the extraction and classification of proper names such as the names of people, locations, and organizations in natural language texts. It is a long-studied topic of natural language processing (NLP) with recent work being mostly on unconventional texts like social media texts such as the study describing a system for named entity recognition on tweets \cite{ritter2011named}. We will develop a named entity recognition system tailored to the energy domain targeting at both the aforementioned basic named entity types and proprietary name classes for the energy domain (such as the names of power plants, feeders, governmental or public energy organizations) which will be able to process both formal and informal texts.
  \item \emph{Temporal expression extraction}: Temporal expressions, such as date and time expressions (particularly when normalized to actual dates), are crucial to pinpoint the exact time of events mentioned in natural language texts. Hence, EneMonIE will include a temporal expression extractor to obtain this information from the media texts. For this purpose, existing state-of-the-art temporal expression extractors such as HeidelTime \cite{strotgen2010heideltime} can be employed and customized accordingly.
  \item \emph{Event extraction}: Extraction of events, regarding security and health, from news articles have long been studied and similar to the case of named entity recognition, recent proposals target informal texts like the study described in \cite{ritter2012open}. Energy-related events occurring all over the world are of public importance. Examples of these events include small or large-scale power outages like blackouts or brownouts that may affect the lives of many people, installation of new power plants with many public or environmental implications, or the invention of a new technology concerning renewable energy production. EneMonIE will have an event extractor particularly tailored to the energy domain to monitor these energy events.
\end{enumerate}

Other functional components of the system will be tools for the following tasks:

\begin{enumerate}
  \item \emph{Text categorization}: This module will be based on the employed energy ontologies and will filter the data sources to determine energy-related textual content. Hence, this categorization stage is the very first operation of EneMonIE on the incoming data stream. The documents passing this filtering stage will then be subject to the information extraction and other data processing stages of the system.
  \item \emph{Social network construction}: This component will be applied to the extracted named entities within the documents to reveal mainly the co-occurrence and coreference relationships between the extracted entities. These social networks can further be employed to determine relationships between the documents as well.
  \item \emph{Sentiment analysis}: Sentiment analysis (or, opinion mining) is a relatively new research topic in NLP which attracts considerable research attention \cite{pang2008opinion}. EneMonIE's sentiment analysis component will be applied to the social media texts to automatically determine the opinions of people on energy-related posts. We believe that this will be an important feature of EneMonIE, as the public opinions on energy events and technologies may affect the related decision-making processes.
  \item \emph{Media interlinking (alignment)}: This component will utilize the findings of the named entity recognition and social media construction modules to interlink different media documents reporting the same topic or event. As it is hard to make sure that this alignment procedure is always highly precise, we will provide percentages denoting the degrees of confidence of this procedure to present these fuzzy results to the users.
  \item \emph{Information fusion and summarization}: This component will combine the results of the aligned documents and provide informative and concise summaries of the related documents, making use of the text summarization techniques both on conventional formal texts like news articles and on informal texts as exemplified with the study on tweets \cite{chakrabarti2011event}. Thereby, the information to be presented to the users are prepared at this stage.
  \item \emph{Web-based information retrieval and visualization}: Finally, Web-based user interface of EneMonIE will facilitate access to the extracted and distilled information regarding energy trends, events, scientific and technological developments, together with the corresponding links to the original data sources whenever applicable. Several significant statistical analysis results regarding this energy information compiled by EneMonIE's functional pipeline will be demonstrated with convenient visualization capabilities.
\end{enumerate}

The system will also comprise ontologies and other resources like lexicons that the system components make use of. There are existing ontologies on topics within the energy domain such as the ontology for electrical power quality \cite{kuccuk2010pqont}, the wind energy ontology \cite{kuccuk2014semi,kuccuk2018ontowind}, and more recently the electrical energy ontology \cite{kuccuk2015high}, to name a few. All of these ontologies are publicly available online for research purposes. Before their prospective employment within EneMonIE, these ontologies will be extended and revised to cover the larger domain of energy. The building processes of the ontologies and lexicons will be a blend of manual engineering and automatic machine learning approaches. EneMonIE will also make use of community-created resources like Wikipedia to enrich its semantic data sources, through automated means. Finally, the system outputs will be stored in a central relational database.\\

EneMonIE will be in operation for media in Turkish and English languages. We plan to ship our first operational prototype system supporting Turkish media. Next, together with the lessons learnt during the building of the first prototype, we will grossly increase the scope of EneMonIE by extending it to be applicable to media in English as well. Nevertheless, we plan to turn it into a multilingual system as a plausible direction of future work.\\

In order to ensure the feasibility of EneMonIE, existing open-source software and libraries for the functional components will be considered for adaptation in the first place. In case we cannot find a viable candidate for a particular task, we will consider building components from scratch through automatic learning, manual engineering, or hybrid methods combining the best practices. The existing knowledge resources like the aforementioned publicly available ontologies and community-created open knowledge bases like Wikipedia will enable us to have a good basis to build our semantic EneMonIE system.

\section{Contributions of EneMonIE}\label{sec:exam}

The main contributions of our Web-based semantic monitoring system for the energy domain, EneMonIE, are listed below:

\begin{itemize}
  \item To the best of our knowledge, this is the first semantic system making heavy use of information processing technologies, particularly those of NLP, information retrieval, and knowledge-based systems, proposed for the high-impact topic of energy. As we have pointed out before, energy is a common asset affecting directly or indirectly the lives of every human being in the world. Therefore, automatic extraction, distillation, fusion and enrichment of energy-related trends, events as well as scientific and technological developments from the Internet and serving this information for public use is a considerable interdisciplinary (broadly combining energy (with related disciplines) and computer science) contribution. It can be considered as a significant accomplishment within the recent research topic of ``energy informatics".
  \item EneMonIE will be a fully-automated system, the components of which will run continuously on the Internet, similar in nature to the open-domain continuous learning systems such as NELL \cite{carlson2010toward}, DeepDive \cite{niu2012deepdive}, and Knowledge Vault \cite{dong2014knowledge} although our system will be a targeted closed-domain one for the energy domain. Hence, the system will continuously extract valuable energy information, then will store this information after associating them with their original sources, and will present the extracted information to the interested parties, including decision-makers, in convenient presentation and visualization formats.
  \item When building the text processing systems for the energy domain, we will pinpoint the challenges of this adaptation and development process. We will discover genuine and novel ways to tackle these challenges, will come up with efficient solutions, and eventually realize these solutions as efficient software systems.
  \item EneMonIE is also significant for its set of diverse data sources, hence can be considered as a multimedia system. Different media types like news articles, news videos, social media texts, blog posts, scholarly articles and technical reports, and open energy data (may be available as linked open data) are all within the scope of the data sources of EneMonIE. Upon the implementation of EneMonIE, diverse documents from these sources will be processed, they will be interlinked, processed to extract important information, sentiments, and social networks, and finally the fused results will be presented for public use. Yet, we should note that only textual data processing is planned for the data sources, therefore, for media like videos, only associated video texts will be processed.
  \item EneMonIE will be a bilingual system, operational on media in Turkish and English languages. But it will also have a modular structure allowing plugging in new components provided that these components comply with the related system interfaces. Therefore, EneMonIE will be an extensible system which can be turned into a multilingual system by equipping it with language-specific text processing tools.
  \item Together with the automatic acquisition of open energy data and information obtained through scholarly articles and technical reports regarding the recent scientific and technological developments on energy, EneMonIE will also act as a hub promoting energy research with its facilities to showcase scientific and technological trends. Interested researchers will be able to carry out deeper analysis of the compiled data and information available through EneMonIE to derive valuable research results.
  \item Wits its diverse data sources, automatic text processing capabilities, and presentation facilities; EneMonIE will be an important source of distilled and concise information for decision-makers including energy generation, transmission, and distribution system operators, energy research centres as well as academicians, students, other individuals interested in the pace of events about energy all over the world.
\end{itemize}

\section{Conclusion}\label{sec:conc}
In this paper, we present the architecture of a semantic system, called EneMonIE, proposed for monitoring energy trends using open data sources available online. EneMonIE will be built as large-scale Web-based semantic system for monitoring up-to-date energy trends through the use of automatic, continuous, and guided information extraction from diverse media available on the Web. Here, with the term \emph{trend}, we mean all energy-related events such as scientific and technological developments; events with social, financial, and environmental impacts; and national and international policy changes. Additionally, EneMonIE will be a hub for open energy information and data published by related energy organizations or researchers. During the building process of EneMonIE, we aim to pinpoint the challenges of the adaptation and development of text processing tools for the energy domain, to come up with novel and efficient solutions to these challenges, and realize these solutions as efficient pieces of software components.

\bibliographystyle{lrec2006}

\begin{thebibliography}{}

\bibitem[\protect\citename{Carlson \bgroup et al.\egroup
  }2010]{carlson2010toward}
Andrew Carlson, Justin Betteridge, Bryan Kisiel, Burr Settles, Estevam~R
  Hruschka, and Tom~M Mitchell.
\newblock 2010.
\newblock Toward an architecture for never-ending language learning.
\newblock In {\em Proceedings of the Twenty-Fourth AAAI conference on
  artificial intelligence}.

\bibitem[\protect\citename{Chakrabarti and Punera}2011]{chakrabarti2011event}
Deepayan Chakrabarti and Kunal Punera.
\newblock 2011.
\newblock Event summarization using tweets.
\newblock In {\em Proceedings of the International AAAI Conference on Web and
  Social Media}, volume~5.

\bibitem[\protect\citename{Dong \bgroup et al.\egroup }2014]{dong2014knowledge}
Xin Dong, Evgeniy Gabrilovich, Geremy Heitz, Wilko Horn, Ni~Lao, Kevin Murphy,
  Thomas Strohmann, Shaohua Sun, and Wei Zhang.
\newblock 2014.
\newblock Knowledge vault: A web-scale approach to probabilistic knowledge
  fusion.
\newblock In {\em Proceedings of the 20th ACM SIGKDD international conference
  on Knowledge discovery and data mining}, pages 601--610.

\bibitem[\protect\citename{K{\"u}{\c{c}}{\"u}k and Arslan}2014]{kuccuk2014semi}
Dilek K{\"u}{\c{c}}{\"u}k and Yusuf Arslan.
\newblock 2014.
\newblock Semi-automatic construction of a domain ontology for wind energy
  using wikipedia articles.
\newblock {\em Renewable Energy}, 62:484--489.

\bibitem[\protect\citename{K{\"u}{\c{c}}{\"u}k and
  K{\"u}{\c{c}}{\"u}k}2018]{kuccuk2018ontowind}
Dilek K{\"u}{\c{c}}{\"u}k and Do{\u{g}}an K{\"u}{\c{c}}{\"u}k.
\newblock 2018.
\newblock Ontowind: An improved and extended wind energy ontology.
\newblock {\em arXiv preprint arXiv:1803.02808}.

\bibitem[\protect\citename{K{\"u}{\c{c}}{\"u}k and
  Yaz{\i}c{\i}}2011]{kuccuk2011exploiting}
Dilek K{\"u}{\c{c}}{\"u}k and Adnan Yaz{\i}c{\i}.
\newblock 2011.
\newblock Exploiting information extraction techniques for automatic semantic
  video indexing with an application to turkish news videos.
\newblock {\em Knowledge-Based Systems}, 24(6):844--857.

\bibitem[\protect\citename{K{\"u}{\c{c}}{\"u}k and
  Yaz{\i}c{\i}}2013]{kuccuk2013semi}
Dilek K{\"u}{\c{c}}{\"u}k and Adnan Yaz{\i}c{\i}.
\newblock 2013.
\newblock A semi-automatic text-based semantic video annotation system for
  turkish facilitating multilingual retrieval.
\newblock {\em Expert systems with applications}, 40(9):3398--3411.

\bibitem[\protect\citename{K{\"u}{\c{c}}{\"u}k \bgroup et al.\egroup
  }2010]{kuccuk2010pqont}
Dilek K{\"u}{\c{c}}{\"u}k, {\"O}zg{\"u}l Salor, Tolga Inan, I{\c{s}}{\i}k
  {\c{C}}ad{\i}rc{\i}, and Muammer Ermi{\c{s}}.
\newblock 2010.
\newblock Pqont: A domain ontology for electrical power quality.
\newblock {\em Advanced Engineering Informatics}, 24(1):84--95.

\bibitem[\protect\citename{K{\"u}{\c{c}}{\"u}k}2015]{kuccuk2015high}
Dilek K{\"u}{\c{c}}{\"u}k.
\newblock 2015.
\newblock A high-level electrical energy ontology with weighted attributes.
\newblock {\em Advanced Engineering Informatics}, 29(3):513--522.

\bibitem[\protect\citename{Niu \bgroup et al.\egroup }2012]{niu2012deepdive}
Feng Niu, Che Zhang, Christopher R{\'e}, and Jude~W Shavlik.
\newblock 2012.
\newblock Deepdive: Web-scale knowledge-base construction using statistical
  learning and inference.
\newblock In {\em Proceedings of the Second International Workshop on Workshop
  on Searching and Integrating New Web Data Sources (VLDS)}.

\bibitem[\protect\citename{Pang and Lee}2008]{pang2008opinion}
Bo~Pang and Lillian Lee.
\newblock 2008.
\newblock Opinion mining and sentiment analysis.
\newblock {\em Foundations and Trends in Information Retrieval}, 2(1-2):1--135.

\bibitem[\protect\citename{Ritter \bgroup et al.\egroup }2011]{ritter2011named}
Alan Ritter, Sam Clark, Oren Etzioni, et~al.
\newblock 2011.
\newblock Named entity recognition in tweets: An experimental study.
\newblock In {\em Proceedings of the Conference on Empirical Methods in Natural
  Language Processing}, pages 1524--1534.

\bibitem[\protect\citename{Ritter \bgroup et al.\egroup }2012]{ritter2012open}
Alan Ritter, Oren Etzioni, and Sam Clark.
\newblock 2012.
\newblock Open domain event extraction from twitter.
\newblock In {\em Proceedings of the 18th ACM SIGKDD International Conference
  on Knowledge Discovery and Data Mining}, pages 1104--1112.

\bibitem[\protect\citename{Str{\"o}tgen and Gertz}2010]{strotgen2010heideltime}
Jannik Str{\"o}tgen and Michael Gertz.
\newblock 2010.
\newblock Heideltime: High quality rule-based extraction and normalization of
  temporal expressions.
\newblock In {\em Proceedings of the 5th International Workshop on Semantic
  Evaluation}, pages 321--324.

\bibitem[\protect\citename{Woon \bgroup et al.\egroup
  }2014]{woon2014forecasting}
Wei~Lee Woon, Zeyar Aung, and Stuart Madnick.
\newblock 2014.
\newblock Forecasting and visualization of renewable energy technologies using
  keyword taxonomies.
\newblock In {\em Proceedings of the International Workshop on Data Analytics
  for Renewable Energy Integration}, pages 122--136. Springer.

\end{thebibliography}

\end{document}